\documentclass{article}
\usepackage{spconf,amsmath,graphicx}
\usepackage{amssymb}
\usepackage{epstopdf}
\usepackage{bbold}
\usepackage{hyperref}
\usepackage{epstopdf}
\usepackage{graphicx}

\usepackage{cite}

\usepackage{graphicx}
\usepackage{caption}
\usepackage{subcaption}
\usepackage{epsfig}
\usepackage{array}
\usepackage{booktabs}
\usepackage{color}
\usepackage{multirow}
\usepackage{url}

\title{Random Feature Maps via a Layered Random Projection (LaRP) Framework for Object Classification}
\name{A. G. Chung, M. J. Shafiee, and A. Wong\thanks{This work was supported by the Natural Sciences and Engineering Research Council of Canada, Ontario Ministry of Economic Development and Innovation  and Canada Research Chairs Program. The authors also thank Nvidia for the GPU hardware used in this study through the Nvidia Hardware Grant Program.}}
\address{Vision \& Image Processing Research Group, System Design Engineering Dept., University of Waterloo\\
\{agchung, mjshafie, a28wong\}@uwaterloo.ca}

\begin{document}

\maketitle

\begin{abstract}
The approximation of nonlinear kernels via linear feature maps has recently gained interest due to their applications in reducing the training and testing time of kernel-based learning algorithms. Current random projection methods avoid the \textit{curse of dimensionality} by embedding the nonlinear feature space into a low dimensional Euclidean space to create nonlinear kernels. We introduce a \textbf{La}yered \textbf{R}andom \textbf{P}rojection (LaRP) framework, where we model the linear kernels and nonlinearity separately for increased training efficiency. The proposed LaRP framework was assessed using the MNIST hand-written digits database and the COIL-100 object database, and showed notable improvement in object classification performance relative to other state-of-the-art random projection methods.

\end{abstract}



\section{Introduction}
\label{Introduction}
The approximation of nonlinear kernels has recently gained popularity due to their ability to implicitly learn nonlinear functions using explicit linear feature spaces~\cite{Scholkopf1999}. These linear feature spaces are typically high (or often infinite) dimensional, and pose what is referred to as the \textit{curse of dimensionality}. To avoid the cost of explicitly working in these feature spaces, the well-known \textit{kernel trick}~\cite{Aizerman1964} is employed where rather than directly learning a classifier in ${\rm I\!R}^d$, a nonlinear mapping $\Phi : {\rm I\!R}^d \rightarrow \mathcal{H}$ is considered such that for all \textbf{x}, \textbf{y} $\in {\rm I\!R}^d$, $\langle \Phi$\textbf{(x)}, $\Phi$\textbf{(y)}$\rangle_\mathcal{H} = K$(\textbf{x}, \textbf{y}) for some kernel $K$(\textbf{x}, \textbf{y}). A classifier $H :$ \textbf{x} $\mapsto$ \textbf{w}$^T \Phi$\textbf{(x)} is then learned for some \textbf{w} $\in \mathcal{H}$.

While this appears to solve the \textit{curse of dimensionality}, it leads to an entirely new problem referred to as the \textit{curse of support}. The support of \textbf{w} can undergo unbounded growth with increasing training data size, resulting in increased training and testing time~\cite{Steinwart2003, Bengio2005}. Though kernel approximation methods have been used successfully in a variety of data analysis tasks, this issue with scalability is becoming more crucial with the advent of big data applications~\cite{Chitta2011, Maji2009}. Initially proposed by Rahimi and Recht~\cite{Rahimi2007}, previous kernel approximation methods have attempted to address the \textit{curse of support} by the low-distortion embedding of the nonlinear feature space $\mathcal{H}$ into a low dimensional Euclidean inner product space via a randomized feature map~\cite{Rahimi2007, Kar2012, Le2013, Pham2013}. 

\begin{figure*}[t]
	\centering
	\includegraphics[width=0.85\textwidth]{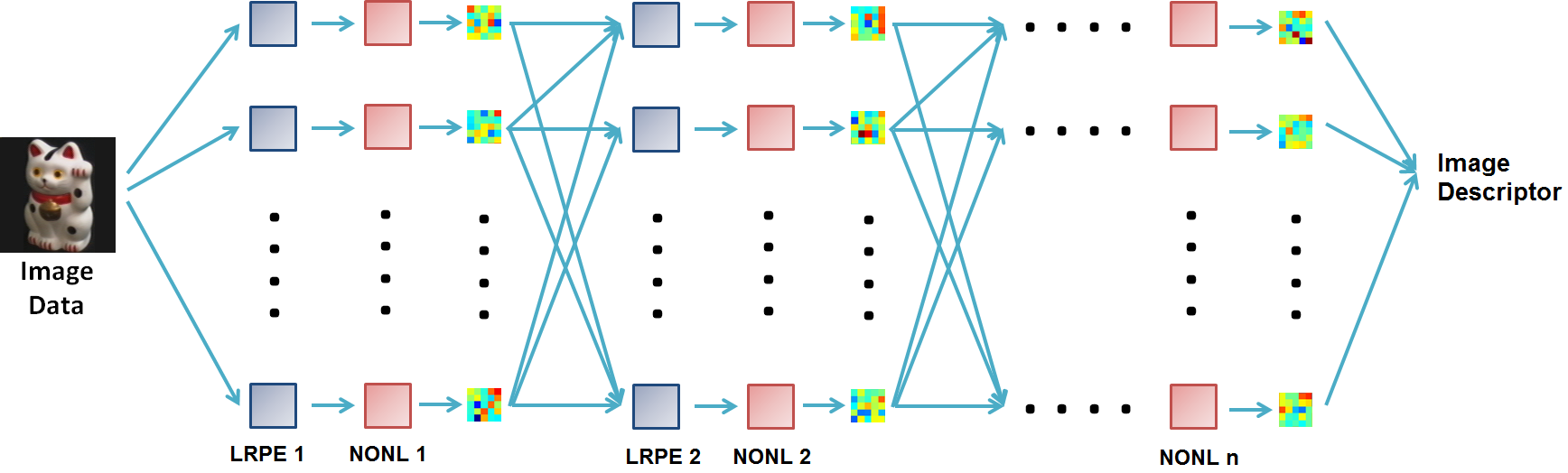}
	\caption{The proposed \textbf{La}yered \textbf{R}andom \textbf{P}rojection (LaRP) framework. The framework is comprised of alternating layers of: i) linear, localized random projection ensembles (LRPE layers), and ii) non-saturating, global nonlinearities (NONL layers) to allow for complex, nonlinear random projections.}
	\label{fig_AlgFrame}
\end{figure*}

Rahimi and Recht~\cite{Rahimi2007} proposed a method for extracting random features via the mapping of the input data to a randomized low-dimensional feature space before applying fast linear methods. Designed to approximate a user specified shift-invariant kernel, Rahimi and Recht evaluated two sets of random features, showing that linear machine learning methods outperformed state-of-the-art large-scale kernel machines in large-scale classification and regression tasks.

Inspired by~\cite{Rahimi2007}, Kar and Karnick~\cite{Kar2012} presented feature maps approximating positive definite dot product kernels via the low-distortion embeddings of dot product kernels into linear Euclidean spaces. Kar and Karnick demonstrated their approach using both homogeneous and non-homogeneous polynomial kernels, as well as the generalization of their approach to compositional kernels. While some experiments resulted in a moderate decrease in classification accuracy, the authors noted that this decrease was almost always accompanied by a significant increase in training and test speeds.

Similarly based on~\cite{Rahimi2007}, Le \textit{et al.}~\cite{Le2013} proposed Fastfood, an approximation method that focuses on significantly decreasing the computational and memory costs associated with kernel methods. Using Hadamard matrices combined with diagonal Gaussian matrices in place of dense Gaussian random matrices, the kernel approximation was shown to have low variance and be unbiased. Le \textit{et al.} achieved similar accuracy to full kernel expansions while being approximately 100x faster and using 1000x less memory.

Pham and Pagh~\cite{Pham2013} presented a novel fast and scalable randomized tensor product technique called \textit{Tensor Sketching} for efficiently approximating polynomial kernels. The images of data were randomly projected without computing the corresponding coordinates in the polynomial feature space, allowing for the fast computation of unbiased estimators. Similar to~\cite{Le2013}, Pham and Pagh demonstrated an increase in accuracy while running orders of magnitude faster than other state-of-the-art methods on large-scale real-world datasets.

More recently, Hamid \textit{et al.}~\cite{Hamid2013} proposed compact random feature maps (CRAFTMaps) as a concise representation of random features maps that accurately approximate polynomial kernels. Taking advantage of the rank deficiency of the spaces constructed by random feature maps, CRAFTMaps achieves this concise representation by up projecting the original data nonlinearly before linearly down projecting the vectors to capture the underlying structure of the random feature space. Hamid \textit{et al.} demonstrated the rank deficiency in the kernel approximations presented by~\cite{Kar2012} and~\cite{Pham2013}, and showed improved test classification errors on multiple datasets by using CRAFTMaps on~\cite{Kar2012} and~\cite{Pham2013} in comparison the original random feature maps.

While these state-of-the-art nonlinear random projection methods have been demonstrated to provide significantly improved accuracy and reduced computational costs on large-scale real-world datasets, they have all primarily focused on embedding nonlinear feature spaces into low dimensional spaces to create nonlinear kernels. As such, alternative strategies for achieving low complexity, nonlinear random projection beyond such kernel methods have not been well-explored, and can have strong potential for improved accuracy and reduced complexity.  

In this work, we propose a novel method for modelling nonlinear kernels using a \textbf{La}yered \textbf{R}andom \textbf{P}rojection (LaRP) framework. Contrary to existing kernel methods, LaRP models nonlinear kernels as alternating layers of linear kernel ensembles and nonlinearities. This strategy allows the proposed LaRP framework to overcome the \textit{curse of dimensionality} while producing more compact and discriminative random features.

\section{Methods}
\label{Methods}
In this paper, we introduce a \textbf{La}yered \textbf{R}andom \textbf{P}rojection (LaRP) framework for object classification. The LaRP framework (shown in Figure~\ref{fig_AlgFrame}) consists of alternating layers of: i) linear, localized random projection ensembles (LRPEs) and ii) non-saturating, global nonlinearities (NONLs). The combination of these layers allows for complex, nonlinear random projections that can produce more discriminative features than can be provided by existing linear random projection approaches.

\subsection{Localized Random Projection Ensemble (LRPE) Layer}
\begin{figure}[h]
	\centering
	\includegraphics[width=\linewidth]{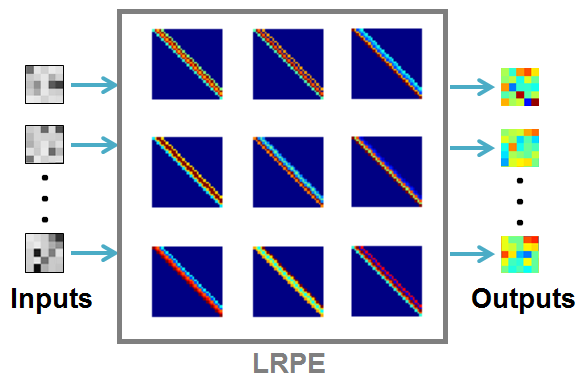}
	\caption{Each localized random projection ensemble (LRPE) sequencing layer consists of an ensemble of $N_i$ localized random projections to project input feature maps from the previous layer to new random feature spaces via banded Toeplitz matrices.}
	\label{fig_SeqUnitCRF}
\end{figure}

Data is projected onto an alternative feature space using random matrices. Each LRPE sequencing layer $i$ (Figure~\ref{fig_SeqUnitCRF}) consists of an ensemble of $N_i$ localized random projections that project input feature maps from the previous layer, \textbf{$X_{k,i-1}$}, to random feature spaces via banded Toeplitz matrices \textbf{$M_{j,i}$}, resulting in output feature maps \textbf{$Y_{j,i}$}:

\begin{equation}
\rm{Y_{j,i} =M_{j,i} X_{k,i-1}},
\end{equation}

\noindent where the banded Toeplitz matrix \textbf{$M_{j,i}$} is a sparse matrix with a support of $n_s$ and can be expressed in the form of:

\begin{equation}
\begin{bmatrix}
K_{j,i} & 0 & 0 & \ldots & 0\\ 
0 & K_{j,i} & 0 & \ldots & 0\\ 
0 & 0 & K_{j,i} & \ldots & 0\\ 
\vdots & & \ddots & & \vdots\\
0 & 0 & \ldots & 0 & K_{j,i}\\ 
\end{bmatrix}
\end{equation}

\noindent \textbf{$M_{j,i}$} can be fully characterized by a random projection kernel $K_{j,i}=[k_0~k_1~\ldots~k_{n_s-1}]$. The sparsity in \textbf{$M_{j,i}$} allows for fast matrix operations and, as a result, facilitates fast localized random projections. 

Each kernel \textbf{$K_{j,i}$} is randomly generated based on a learned distribution $P_{j,i}$. In this work, $P_{j,i}$ is a uniform distribution where the upper and lower bounds of the distribution, $[a, b]$, are learned during the training process. 

\subsection{Nonlinearity (NONL) Layer}
The output feature maps from a LRPE layer is fed into a nonlinearity (NONL) layer. Each NONL layer consists of an absolute value rectification (AVR) followed by a sliding-window median regularization (SMR). 

\begin{figure}[h]
	\centering
	\includegraphics[width=\linewidth]{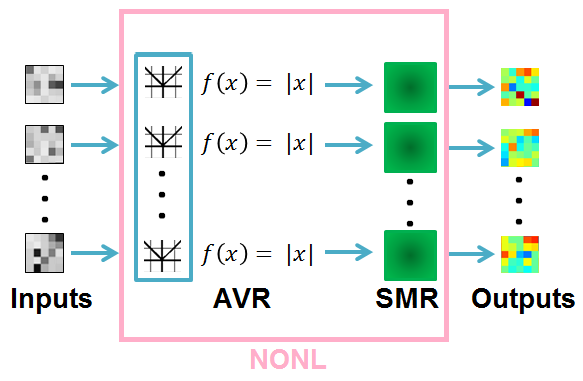}
	\caption{Each nonlinearity (NONL) layer consists of an absolute value rectification (AVR) to introduce non-saturating nonlinearity into the layered random projection framework, followed by sliding-window median regularization (SMR) to improve robustness to uncertainties in the data.}
	\label{fig_SeqUnit}
\end{figure}

An absolute value rectification (AVR) is first applied to an output feature map from the preceding LRPE sequencing layer (i.e., $Y_{j,i}$) to produced the rectified feature map $Y_{j,i}^{\rm rect}$, and can be defined as follows:

\begin{equation}
	Y_{j,i}^{\rm rect} = |Y_{j,i}|.
\end{equation}

\noindent The purpose of AVR is to introduce non-saturating nonlinearity into the input feature map to allow for complex, non-linear random projections.  

The AVR is followed by a sliding-window median regularization (SMR) applied to the rectified feature map $Y_{j,i}^{\rm rect}$. This SMR produces the final output of a pair of LRPE and NONL layers $X_{j,i}$, and can be defined as

 \begin{equation}
 X_{j,i}(x,y) = Median(Y_{j,i}^{\rm rect}(x,y)|R)
  \end{equation}
  
\noindent where $R$ is a sliding window, and is used to nonlinearly enforce spatial consistency within the rectified feature map $Y_{j,i}^{\rm rect}$. A 3$\times$3 sliding window is used in this study as it was empirically shown to provide a good balance between spatial consistency and feature map information preservation. 

\subsection{Training LaRP}
Each random projection kernel \textbf{$K_{j,i}$} characterizing the random projection matrices in the proposed LaRP framework has two parameters that needs to be trained: the upper and lower bounds of the uniform distribution, $[a, b]$. The total number of parameters to be trained in the LaRP framework is $\sum_{i=1}^{N_l} 2N_i$, where $N_l$ is the number of layers. In this work, the LaRP framework is trained via iterative scaled conjugate gradient optimization using cross-entropy as the objective function.

\section{Results}
\label{Results}
\begin{table*}[t!]
	\centering
	\caption{Test classification errors for MNIST and COIL-100 databases, with the best classification error for each database in boldface. The proposed LaRP framework was compared against Fastfood~\cite{Le2013}, RKS~\cite{Rahimi2007}, RFM~\cite{Kar2012}, TS~\cite{Pham2013}, and CM applied to RFM and TS~\cite{Hamid2013}. Note that while Fastfood, RKS, RFM, TS, and CM used $2^{12}$ and $2^{15}$ features, the proposed LaRP framework used $2^{10}$ random features.}
	\begin{tabular}{|l|c|c|c|c|c|c|c|c|c|c|c|c|c|}
	\hline
	\multirow{2}{*}{} & \multicolumn{2}{c|}{Fastfood~\cite{Le2013}} & \multicolumn{2}{c|}{RKS~\cite{Rahimi2007}} & \multicolumn{2}{c|}{RFM~\cite{Kar2012}} & \multicolumn{2}{c|}{TS~\cite{Pham2013}} & \multicolumn{2}{c|}{CM RFM~\cite{Hamid2013}} & \multicolumn{2}{c|}{CM TS~\cite{Hamid2013}} & LaRP \\ \cline{2-14}
	& \textbf{$2^{12}$} & \textbf{$2^{15}$} & \textbf{$2^{12}$} & \textbf{$2^{15}$} & \textbf{$2^{12}$} & \textbf{$2^{15}$} & \textbf{$2^{12}$} & \textbf{$2^{15}$} & \textbf{$2^{12}$} & \textbf{$2^{15}$} & \textbf{$2^{12}$} & \textbf{$2^{15}$} &  \textbf{$2^{10}$} \\
	\hline
	\textbf{MNIST~\cite{Lecun1998}}	&	$2.78$ 	&	$1.87$	&	$2.94$ 	&	$1.91$	
&	$3.17$ 	&	$1.62$	&	$3.25$ 	&	$1.65$	&	$3.09$ 	&	$1.52$	&	$2.90$ 	&	$1.44$	&\textbf{1.30} \\ \hline
	\textbf{COIL-100~\cite{Nene1996}} &	$7.83$ 	&	$5.21$	&	$7.36$ 	&	$4.81$	&	$7.55$ 	&	$4.83$ 	&	$7.19$ 	&	$4.27$ 	&	$6.86$ 	&	$4.08$ 	&	$5.97$ 	&	$3.96$	&\textbf{0.36} \\ \hline
	\end{tabular}
	\label{tab_Results}
\end{table*}

\subsection{Experimental Setup}
\begin{figure}[h]
	\begin{center}
	\begin{tabular}{cccccc}
		\includegraphics[scale = 1]{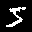}	\hspace*{-5pt}&
		\includegraphics[scale = 1]{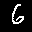}	\hspace*{-5pt}&
		\includegraphics[scale = 1]{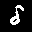} \hspace*{-5pt}&
		\includegraphics[scale = 1]{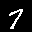} \hspace*{-5pt}&
		\includegraphics[scale = 1]{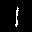} \hspace*{-5pt}&
		\includegraphics[scale = 1]{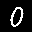} \\
		\includegraphics[scale = 0.25]{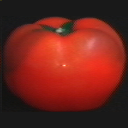} \hspace*{-5pt}&
		\includegraphics[scale = 0.25]{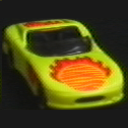} \hspace*{-5pt}&
		\includegraphics[scale = 0.25]{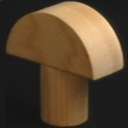} \hspace*{-5pt}&
		\includegraphics[scale = 0.25]{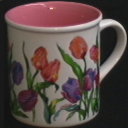} \hspace*{-5pt}&
		\includegraphics[scale = 0.25]{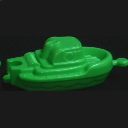} \hspace*{-5pt}&
		\includegraphics[scale = 0.25]{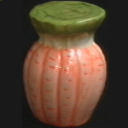}	
	\end{tabular}
	\vspace{-10pt}
	\end{center}
	\caption{Sample images from the databases used for testing. The first row shows sample images from the MNIST hand-written digits database, and the second row shows samples from the COIL-100 object database.}
	\label{fig_Data}
\end{figure}

To assess the efficacy of the proposed LaRP framework for object classification, our method was compared to state-of-the-art random projection methods~\cite{Le2013, Rahimi2007, Kar2012, Pham2013, Hamid2013} via the MNIST hand-written digits database~\cite{Lecun1998} and the COIL-100 object database~\cite{Nene1996}. The MNIST database was divided into training and testing sets as specified by~\cite{Lecun1998}. Similar to~\cite{Pham2006} and~\cite{Murase1995}, the COIL-100 database was divided into two equally sized partitions for training and testing; the training set consisted of 36 views of each object at $10^{\circ}$ intervals, and the remaining views were used for testing. Figure~\ref{fig_Data} shows sample images from the MNIST hand-written digits database (first row) and the COIL-100 object database (second row).

Table~\ref{tab_Kernels} summarizes the number of localized random projections and the supports of the random projection matrices used in each LRPE sequencing layer for the proposed LaRP framework in this study.

\begin{table}[h]
	\centering
	\caption{Summary of number of localized random projections and supports of the random projection matrices at each LRPE sequencing layer for the LaRP framework used in this study.}
	\begin{tabular}{>{\centering\arraybackslash}m{2.5cm}|>{\centering\arraybackslash}m{2.5cm}|>{\centering\arraybackslash}m{2.5cm}}
		\hline
		\textbf{LRPE Sequencing Layer} & \textbf{Number of Projections} & \textbf{Projection Matrix Support ($n_s$)} \\
		\hline
		1	& 	$512$  		&	$25$	\\
		2	& 	$512$  		&	$25$	\\
		3	& 	$1024$  	&	$25$	\\
		\hline
	\end{tabular}
	\label{tab_Kernels}
\end{table}	

The proposed LaRP framework characterizes a given image using $2^{10}$ random features. The test classification errors for the state-of-the-art methods were obtained from~\cite{Hamid2013}. As the test classification errors for the state-of-the-art methods were unavailable for $2^{10}$ features, the classification errors of the state-of-the-art methods using $2^{12}$ and $2^{15}$ features were used for comparison.

\subsection{Experimental Results}
Table~\ref{tab_Results} shows the test classification errors for the proposed LaRP framework and state-of-the-art methods Fastfood~\cite{Le2013}, RKS~\cite{Rahimi2007}, RFM~\cite{Kar2012}, TS~\cite{Pham2013}, and CM applied to RFM and TS~\cite{Hamid2013} using the MNIST and COIL-100 databases. The test classification errors clearly indicate that the proposed LaRP framework outperformed the state-of-the-art methods.

The proposed LaRP framework achieved test classification error improvements of 1.48 and 0.14 over the best state-of-the-art results using $2^{12}$ and $2^{15}$ features, respectively, for the MNIST database. Similarly, LaRP outperformed the state-of-the-art methods using the COIL-100 database, showing test classification error improvements of 5.61 and 3.60 over the best state-of-the-art results using $2^{12}$ and $2^{15}$ features, respectively.

In addition to the notable test classification error improvements, it can also be observed that the proposed LaRP framework achieved these results using significantly fewer random features. While the state-of-the-art methods require $2^{15}$ features to generate comparable test classification errors, the LaRP framework attained better test classification errors using $2^{10}$ features. This indicates that the LaRP framework is capable of generating more compact and discriminative random features relative to state-of-the-art methods.

\section{Conclusion}
\label{Conclusion}
A novel \textbf{La}yered \textbf{R}andom \textbf{P}rojection (LaRP) framework was presented, where we overcome the \textit{curse of dimensionality} and model the linear kernels and nonlinearity separately. This is done via alternating layers of i) linear, localized random projection ensembles (LRPE layers), and ii) non-saturating, global nonlinearities (NONL layers) to allow for complex, nonlinear random projections.

The proposed LaRP framework was evaluated against state-of-the-art random kernel approximation methods using the MNIST and COIL-100 databases. Generating $2^{10}$ random features, the LaRP framework achieved the lowest test classification errors for both databases (1.30 for MNIST, 0.36 for COIL-100) when compared to the state-of-the-art methods using $2^{12}$ and $2^{15}$ random features. This indicates the potential of the proposed LaRP framework for producing useful and compact feature maps for object classification.

Future work includes the investigation of inter-kernel information and its effects on the generated random features maps. As well, comprehensive evaluation and investigation of the LaRP framework for different image recognition and processing tasks and applications, e.g., saliency, segmentation, etc., will be conducted.

\bibliographystyle{IEEEtran}
%
%
%

\bibliography{RandomNets}

%

%






\end{document}